\documentclass[conference]{IEEEtran}
\IEEEoverridecommandlockouts
\usepackage{cite}
\usepackage{amsmath,amssymb,amsfonts}
\usepackage{algorithmic}
\usepackage{graphicx}
\usepackage{textcomp}
\usepackage{xcolor}
\usepackage{booktabs}
\usepackage{color}
\usepackage{enumitem}
\def\BibTeX{{\rm B\kern-.05em{\sc i\kern-.025em b}\kern-.08em
    T\kern-.1667em\lower.7ex\hbox{E}\kern-.125emX}}
\begin{document}

\title{Cheap-fake Detection with LLM using Prompt Engineering
\thanks{This work is partially supported by the Sichuan Provincial Key Laboratory of Intelligent Terminals (Grant Number: SCITLAB-20016). Corresponding authors are Xiaohong Liu (xiaohongliu@sjtu.edu.cn) and Wenyi Wang (wangwenyi@uestc.edu.cn).}
}

\author{
\IEEEauthorblockN{Guangyang Wu}
\IEEEauthorblockA{
\textit{University of Electronic Science}\\ \textit{and Technology of China}\\
Chengdu, China 
}
\and
\IEEEauthorblockN{Weijie Wu}
\IEEEauthorblockA{
\textit{University of Electronic Science}\\ \textit{and Technology of China}\\
Chengdu, China}
\and
\IEEEauthorblockN{Xiaohong Liu}
\IEEEauthorblockA{
\textit{Shanghai Jiao Tong University}\\
Shanghai, China}
\and
\IEEEauthorblockN{Kele Xu}
\IEEEauthorblockA{
\textit{National University of Defense Technology} \\
Changsha, China}
\and
\IEEEauthorblockN{Tianjiao Wan}
\IEEEauthorblockA{
\textit{National University of Defense Technology} \\
Changsha, China}
\and
\IEEEauthorblockN{Wenyi Wang}
\IEEEauthorblockA{
\textit{University of Electronic Science}\\ \textit{and Technology of China}\\
Chengdu, China}
}

\maketitle

\begin{abstract}
The misuse of real photographs with conflicting image captions in news items is an example of the out-of-context (OOC) misuse of media. In order to detect OOC media, individuals must determine the accuracy of the statement and evaluate whether the triplet (~\textit{i.e.}, the image and two captions) relates to the same event. This paper presents a novel learnable approach for detecting OOC media in ICME'23 Grand Challenge on Detecting Cheapfakes. The proposed method is based on the COSMOS structure, which assesses the coherence between an image and captions, as well as between two captions. We enhance the baseline algorithm by incorporating a Large Language Model (LLM), GPT3.5, as a feature extractor. Specifically, we propose an innovative approach to feature extraction utilizing prompt engineering to develop a robust and reliable feature extractor with GPT3.5 model. The proposed method captures the correlation between two captions and effectively integrates this module into the COSMOS baseline model, which allows for a deeper understanding of the relationship between captions. By incorporating this module, we demonstrate the potential for significant improvements in cheap-fakes detection performance. The proposed methodology holds promising implications for various applications such as natural language processing, image captioning, and text-to-image synthesis. Docker for submission is available at https://hub.docker.com/repository/docker/mulns/ acmmmcheapfakes.

\end{abstract}


\section{Introduction}
\label{sec:intro}
The emergence of social media has dramatically altered the process by which people access information, resulting in a significant upswing in available information, hastening the spread of fake news and other types of misleading data. On social media, two principal forms of deceptive information regularly encountered are deepfakes and cheapfakes~\cite{cosmos}. While "deepfakes" refer to videos that have been doctored using machine learning or other AI-based methods to create or combine human faces and bodies, "cheap fakes" are a far less costly variant of doctored videos generated through readily available software tools such as Photoshop, PremierePro, among others. Cheapfakes are often created through manipulating image captions, image editing or by adjusting video speed.
Due to their ease of creation, cheapfakes are perceived to be more ubiquitous and damaging than deepfakes. The out-of-context use of images, where unaltered photos are put to new and deceptive contexts, is one of the main reasons cheapfakes pose such a danger. This phenomenon occurs when an image is sourced from various locations with contradictory or conflicting captions. The detection of misinformation based on out-of-context images is particularly arduous since the visual content is unchanged, and the misleading or incorrect information is only conveyed through the image-text combination.

The COSMOS baseline framework~\cite{cosmos} employs a two-step method to address the problem of detecting cheap-fake images. The first step involves an image-text matching module that evaluates the coherence between an image and its caption. The second step utilizes an out-of-context (OOC) detection module to predict the final outcome. The approach relies on semantic textual similarity (STS) scores to determine whether an image-caption pair is OOC or not (NOOC). S-BERT is used to calculate the semantic similarity between two captions, where the input is a pair of captions and the output is a similarity score in the range from 0 to 1. If the similarity score is less than the pre-defined threshold, the triplet is predicted as out-of-context.

However, there are certain scenarios where the STS scores may not perform well. For instance, if two captions are contradictory, the STS model may produce a high score if the ratio of similar words is high. Conversely, a pair of entailment captions may have a low STS score if one caption is much more detailed than the other. Therefore, a more comprehensive evaluation of the relationship between two captions is critical. Tran~\textit{et al.}~\cite{texture} proposed using a Natural Language Inference (NLI) model to determine whether the given "hypothesis" and "premise" logically follow (entailment) or unfollow (contradiction) or are undetermined (neutral) with each other. However, the performance of this model is still limited and may produce unreliable results for challenging cases. 
In recent years, Large Language Models (LLMs) have emerged, which possess enhanced semantic comprehension abilities compared to conventional BERT-based models. This development has motivated the use of LLM as a powerful tool for evaluating the coherence between two captions. To enhance detection accuracy in cases where previous methods may not be effective, we leverage the GPT3.5 model. This model has shown remarkable results in various NLP tasks and can provide a more comprehensive evaluation of the relationship between two captions. By utilizing the GPT3.5 model, we can address the limitations of previous methods and enhance the overall performance of the COSMOS framework. Nevertheless, two key challenges must be addressed. Firstly, the full parameters of GPT3.5 is not currently openly accessible, and its usage is limited to the OpenAI API. Secondly, the adaptive nature of GPT3.5 and its frequent updates can result in dynamic and potentially unstable outcomes.

This study presents an innovative approach to feature extraction utilizing prompt engineering to develop a robust and reliable feature extractor. The proposed method captures the correlation between two captions and effectively integrates this module into the COSMOS baseline model. Our study emphasizes the significance of prompt engineering in feature extraction, which allows for a deeper understanding of the relationship between captions. By incorporating this method into the baseline model, we demonstrate the potential for significant improvements in cheap-fakes detection performance. The proposed methodology holds promising implications for various applications such as natural language processing, image captioning, and text-to-image synthesis. 

\begin{figure*}[ht]
  \centering
  \includegraphics[width=0.86\linewidth]{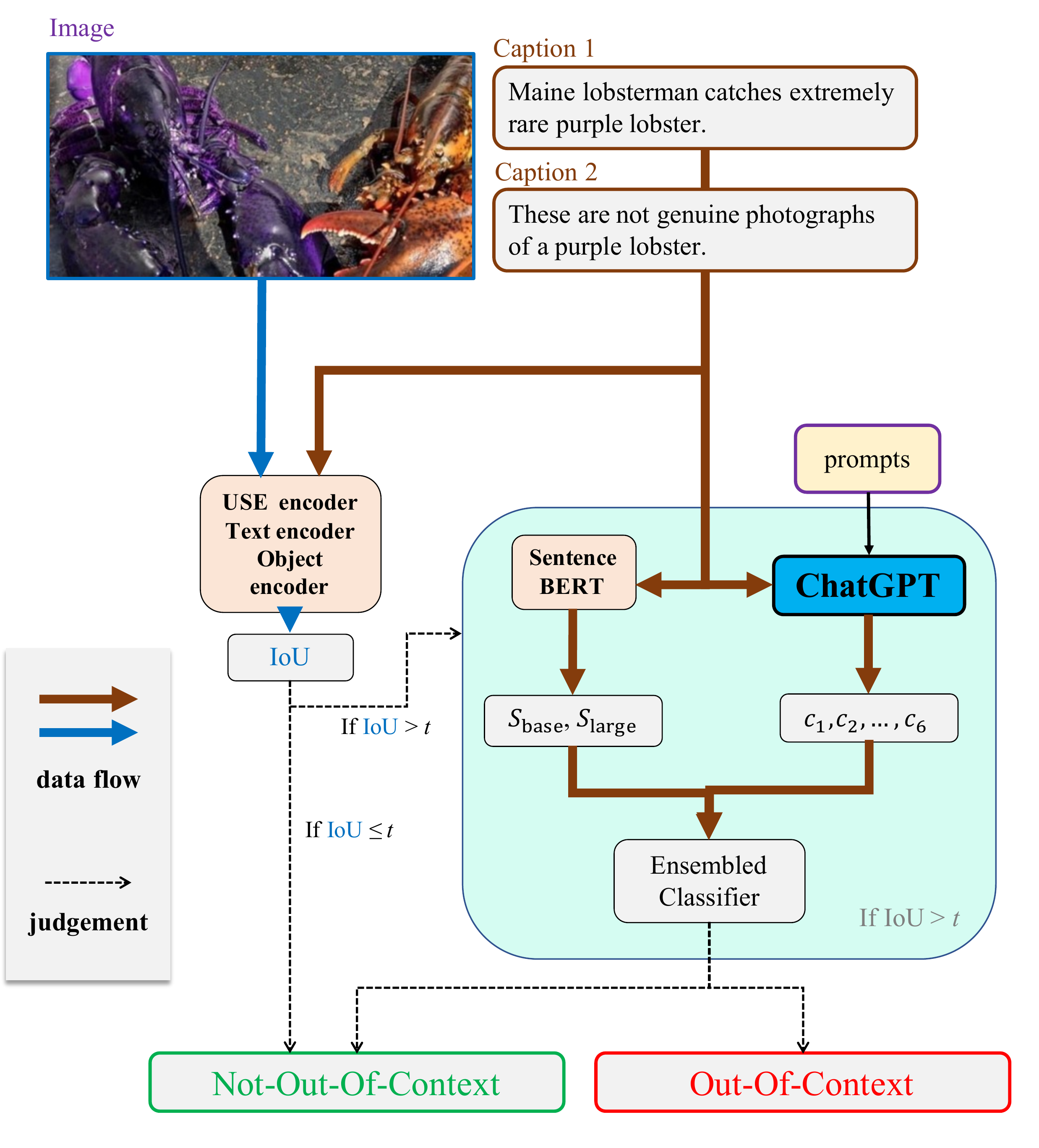}
  \caption{Overall framework of our method.}
  \label{fig:main}
\end{figure*}

\section{Related work}

\subsection{Cheapfakes Detection}
The baseline method, COSMOS~\cite{cosmos}, utilizes image-text matching and a BERT-based module to detect cheap-fakes. Akgul~\textit{et al.}~\cite{cosmos} have generated a dataset of 200,000 images with 450,000 textual captions to train the image-text matching model. The COSMOS use a heuristic pipeline to determine whether an image-caption triplet is out-of-context (OOC).
In the "ACMMM 2022 Grand Challenge on Detecting Cheapfakes", La~\textit{et al.}~\cite{multimodal} proposed to use image captioning method to enhance accuracy~\cite{multimodal}. They convert the image to caption using a image captioning model,
and extract the features with the RoBERTa model. Afterwards, they use an ANN or SVM for binary classifying according to features of three captions. Moreover, La~\textit{et al.}~\cite{combined} proposed a Visual Semantic Reasoning (VSRN) method to enhance the image-text matching module. They use a pre-trained DeBERTa model to obtain additional semantic information between two captions.  Tran~\textit{et al.}~\cite{texture} proposed using a Natural Language Inference (NLI) model to determine whether the given "hypothesis" and "premise" logically follow (entailment) or unfollow (contradiction) or are undetermined (neutral) with each other. Furthermore, they use online-search to address the hard cases for image-caption matching.

\subsection{Fact-Checking}
Fact-checking is a vital task aimed at evaluating the veracity of statements made by prominent individuals, including politicians, pundits, and other public figures \cite{vlachos-riedel-2014-fact,oshikawa2020survey}. In recent years, researchers have proposed various techniques to address the problem of fact-checking. Shi~\textit{et al.}~\cite{path_mining} proposed a discriminative path-based approach for fact-checking in knowledge graphs. The method incorporates connectivity, type information, and predicate interactions to identify true statements accurately. Jin~\textit{et al.}~\cite{att-RNN} introduced a novel Recurrent Neural Network (RNN) with an attention mechanism (att-RNN) to integrate multimodal features for rumor detection effectively. Ferreira~\textit{et al.}~\cite{stance-detection} introduced the Emergent dataset, which is a digital journalism project aimed at debunking rumors. This dataset has been used to develop effective stance detection techniques.

\section{Methods}

This section presents a novel approach for detecting cheapfakes that incorporates both textual and visual information using a learnable multimodal method. We build upon the baseline algorithm of the challenge~\cite{challenge}, COSMOS, which is a two-step approach that evaluates the coherence between an image and two captions, followed by the coherence between the two captions themselves. Our proposed method enhances the baseline by introducing a feature extractor, built using a Large Language Model (LLM) called GPT3.5, to improve detection accuracy. The method is designed to solve the first task of the ICME'23 Grand Challenge~\cite{challenge} on Detecting Cheapfakes, which focuses on detecting whether an image-caption triplet is in or out of context. Figure~\ref{fig:main} depicts an overview of our proposed method.

\subsection{Coherence between Image and Captions}
Our method utilize the Image-text matching module, which was introduced in a prior work referred to as COSMOS~\cite{cosmos}. This module is designed to take as input an image-caption pair, and to generate a bounding box with an image-caption coherence score. A higher score value indicates a greater level of coherence between the input image and caption. The bounding box produced by this module frames the region that corresponds to the maximum coherence score.

The Image-text matching module operates by first taking the caption as input, which is then processed by the Universal Sentence Encoder (USE) to produce a 512-dimensional vector. This vector is then further processed by the Text Encoder to obtain a 300-dimensional vector. Concurrently, the image is inputted into a pre-trained Mask-RCNN to generate up to 10 object bounding boxes. For each object, the Object Encoder is used to produce a 300-dimensional embedding vector.
To evaluate the coherence between the image and caption, the intersection-over-union (IoU) score is calculated through a dot product of embedding vectors. If the value of the IoU score is lower than a pre-defined threshold, the triplet is predicted as NOOC. In contrast, if the IoU score value is greater than the threshold, further estimation is carried out for the captions in the triplet. In this work, we set the threshold value to 0.25, which is determined based on prior studies~\cite{texture}.

\subsection{Coherence between captions}
Once the threshold is met, the proposed method evaluates the textual semantic relationship between the two captions in the triplet for further classification. In addition to the similarity of the two sentences estimated by BERT-based models, we introduce a feature vector that represents the coherence of the two sentences. Specifically, we first estimate a similarity vector $[S_{base}, S_{large}]$ using SBERT and BERT-large models for each pair of captions following~\cite{cosmos}, which represents the semantic similarity between them.

Next, we utilize the GPT3.5 model (also denoted as ChatGPT) to generate a discriminated vector $[c_1, c_2, \dots, c_6]$ that represents the semantic relation between the captions regarding various features, including the probability of being out of context, the consistency of subject matters, the consistency of broader context, coherence, information completeness, and semantic similarity. In order to guide the GPT3.5 model to conclude these features, we carefully designed prompts as shown below:

``Given two sentences, I am going to ask you six questions. You should provide a final answer in a python list of length 6 where each component is a rate value (integer ranging from 0 to 9).
\begin{itemize}[noitemsep, leftmargin=*]
\item The first question: Determine whether these two sentences are out of context. Rate your judgment by an integer number ranging from 0 to 9, where 9 refers to being completely out of context, and 0 refers to being completely in context.
\item The second question: Determine whether the subject matters of these two sentences are the same. Rate your judgment by an integer number ranging from 0 to 9, where 9 indicates that the subject matters are completely different, and 0 indicates that the subject matters are completely the same
\item The third question: Determine whether the broader context of these two sentences refer to are the same. Rate your judgment by an integer number ranging from 0 to 9, where 9 indicates that the broader context is completely different, and 0 indicates that the broader context is completely the same
\item The fourth question: Determine whether these two sentences cohere together. Please rate your judgment by an integer number ranging from 0 to 9, where 9 indicates that the two sentences are not coherent at all, and 0 indicates that the two sentences are highly coherent
\item The fifth question: Determine whether any information is missing that could help to explain the relationship between the two sentences. Please rate your judgment by an integer number ranging from 0 to 9, where 9 indicates that  important information is missing, and 0 indicates that there is no information missing.
\item The sixth question: Determine the semantic similarity between the two sentences. Semantic similarity should be rated by an integer number ranges from 0 to 9, where 0 refers to semantically identical, and 9 refers to completely semantic different.
\end{itemize}
The two sentences are [CAPTION1, CAPTION2]. You should output the python list only without explanations.''

To produce a stable output, we leverage the OpenAI API to implement the feature extractor using the pre-trained model `gpt-3.5-turbo-0301'. Specifically, this model is a snapshot of `gpt-3.5-turbo' from March 1st 2023. Unlike `gpt-3.5-turbo', this model will not receive updates, and will only be supported for a three month period ending on June 1st 2023. Furthermore, the temperature value also effect the randomness of the model. Specifically, the higher temperature make the output more random, while lower values will make it more focused. Therefore, we set the temperature to 0 to obtain deterministic results. Once the similarity vector $[s_{base}, s_{large}]$ and the GPT vector $[c_1, \dots, c_6]$ are computed for each input sample, they are passed as input to an ensemble classifier, AdaBoost~\cite{adaboost}. The AdaBoost classifier is a meta-estimator that begins by fitting a classifier on the original dataset and then fits additional copies of the classifier on the same dataset but where the weights of incorrectly classified instances are adjusted such that subsequent classifiers focus more on difficult cases. This classifier utilizes the extracted features to generate predictions of either 0 for NOOC or 1 for OOC.

\section{Experiments}
In this section, we describe the training details and the testing results compared with previous methods.
\subsection{Training Details}
To train the binary classifier for predicting the OOC or NOOC label based on the feature vector, we partitioned the public testing dataset of the ICME'23 Grand Challenge on Detecting Cheapfakes into a training set (50\%) and a testing set (50\%). The public testing dataset comprised 1000 samples, each of which consisted of an image and two captions as inputs, along with the corresponding OOC or NOOC labels. Due to the limited number of samples available for training, we adopted a set of simple classifiers to prevent overfitting while achieving superior performance. We trained the classifiers using the 5-fold cross-validation method.
\subsection{Experimental Results}
Table~\ref{tab:main} presents a comparison of the results obtained from various classifiers, including Support Vector Machine (SVM), Random Forest (RF), and AdaBoost, with those obtained from previous methods. The second column of the table indicates the accuracy on the testing dataset (i.e., half of the public test dataset), while the third column represents the accuracy on the entire public test dataset. The GPT+AdaBoost classifier was selected as the final solution since it achieved the highest score on the testing dataset and demonstrated better generalization ability than GPT+RF.

\begin{table}[htbp]
  \centering
  \caption{Comparisons with other methods. We emphasize the best accuracy with \textbf{bold} text.}
  \begin{tabular}{ccc}
    \toprule
    Methods & Acc. (test) & Acc. (full) \\
    \midrule
    COSMOS~\cite{cosmos} & 0.839 & 0.821 \\
    Tran~\textit{et al.}~\cite{texture} & - & 0.891 \\
    La~\textit{et al.}~\cite{multimodal} & - & 0.867 \\
    La~\textit{et al.}~\cite{combined} & - & 0.760 \\
    GPT+SVM (ours) & 0.852 & 0.867 \\
    GPT+RF (ours) & 0.858 & \textbf{0.929} \\
    GPT+AdaBoost (ours) & \textbf{0.894} & 0.888 \\
    \bottomrule
  \end{tabular}

  \label{tab:main}
\end{table}

\section{Conclusion}
In conclusion, our proposed method uses the COSMOS structure to evaluate the coherence between an image and two captions, and between two captions. The method employs first estimate the image-caption coherence represented by an IoU value, and S-BERT and BERT-large models to estimate a similarity vector. Additionally, we use the GPT-3.5 model to generate a discriminative vector that represents the semantic relation between the captions based on a set of carefully designed features. The use of these methods and models results in an accurate and stable representation of the coherence between image-caption triplets, and improve the baseline model by large margin.

\clearpage
\newpage
\bibliographystyle{IEEEbib}
\bibliography{conference_101719}

\end{document}